\documentclass[conference]{IEEEtran}
\IEEEoverridecommandlockouts
\usepackage{cite}
\usepackage{amsmath,amssymb,amsfonts}
\usepackage{algorithmic}
\usepackage{graphicx}
\usepackage{textcomp}
\usepackage{xcolor}
\def\BibTeX{{\rm B\kern-.05em{\sc i\kern-.025em b}\kern-.08em
    T\kern-.1667em\lower.7ex\hbox{E}\kern-.125emX}}

\usepackage{amssymb}
\usepackage{pifont}
\usepackage{makecell}
\usepackage[para,online,flushleft]{threeparttable}
\usepackage{tikz}

\newcommand{\cmark}{\ding{51}}%
\newcommand{\xmark}{\ding{55}}%

\begin{document}

\newcommand\copyrighttext{\footnotesize \textcopyright~2023 IEEE. Personal use of this material is permitted.  Permission from IEEE must be obtained for all other uses, in any current or future media, including reprinting/republishing this material for advertising or promotional purposes, creating new collective works, for resale or redistribution to servers or lists, or reuse of any copyrighted component of this work in other works.
}%

\title{Weakly Supervised Multi-Modal 3D Human Body Pose Estimation for Autonomous Driving\\

%

}



\author{
Peter Bauer$^{1}$, Arij Bouazizi$^{2}$, Ulrich Kressel$^{3}$, Fabian B. Flohr$^{4}$
\thanks{$^{1}$University of Stuttgart, Keplerstraße 7, 70174 Stuttgart (Contact Author: st148923@stud.uni-stuttgart.de)}
\thanks{$^{2}$Friedrich-Alexander-Universität Erlangen-Nuernberg, Cauerstraße 7, 91058 Erlangen, Germany}
\thanks{$^{3}$Ulm University, Albert-Einstein-Allee 41, 89081, Ulm, Germany.}
\thanks{$^{4}$Munich University of Applied Sciences, Intelligent Vehicles Lab, Lothstraße 34, 80335, Munich, Germany.}
\thanks{E-mail: \textit{firstname.lastname@\{fau.de, uni-ulm.de, hm.edu\}}.}%

}

\newcommand\copyrightnotice{%
	\begin{tikzpicture}[remember picture,overlay]
	\node[anchor=south,xshift=0pt,yshift=14pt] at (current page.south) {\fbox{\parbox{\dimexpr\textwidth-\fboxsep-\fboxrule\relax}{\copyrighttext}}};
	\end{tikzpicture}%
}

\maketitle

\begin{abstract}

Accurate 3D human pose estimation (3D HPE) is crucial for enabling autonomous vehicles (AVs) to make informed decisions and respond proactively in critical road scenarios. Promising results of 3D HPE have been gained in several domains such as human-computer interaction, robotics, sports and medical analytics, often based on data collected in well-controlled laboratory environments. Nevertheless, the transfer of 3D HPE methods to AVs has received limited research attention, due to the challenges posed by obtaining accurate 3D pose annotations and the limited suitability of data from other domains.

We present a simple yet efficient weakly supervised approach for 3D HPE in the AV context by employing a high-level sensor fusion between camera and LiDAR data. The weakly supervised setting enables training on the target datasets without any 2D~/~3D keypoint labels by using an off-the-shelf 2D joint extractor and pseudo labels generated from LiDAR to image projections. Our approach outperforms state-of-the-art results by up to $\sim$ 13\% on the Waymo Open Dataset in the weakly supervised setting and achieves state-of-the-art results in the supervised setting.

\end{abstract}


\section{Introduction}

Human Pose Estimation (HPE) is known as a fundamental problem in computer vision with various applications including autonomous vehicles \cite{wiederer2020traffic,czech2022board}, human-robot interaction and augmented reality. It is defined as the localization of specific body joints in 2D~/~3D space and provides a simple encoding with respect to a reference coordinate system. Current research exploits various ways of solving the task of HPE in the 2D image space \cite{fang2022alphapose} as well as in the 3D domain \cite{Bouazizi2021,bouazizi2021learning}. While approaches that operate in the 2D domain mainly use RGB data, 3D HPE algorithms need to incorporate additional depth information or multiple viewpoint calibrated and time synchronized camera systems to overcome the projection ambiguity. Although many methods have been proposed in the literature, there is a shortage in the field of autonomous driving, mainly caused by the fact that obtaining accurate 3D pose annotations is time-consuming and expensive for uncontrolled outdoor scenarios.
However, to ensure safe operation in complex urban environments, it is critical for autonomous vehicles to have the ability to understand and interpret human posture. This enables the estimation of the intentions of vulnerable road

\copyrightnotice

\begin{figure}[htbp]
\centerline{\includegraphics[width=0.45\textwidth]{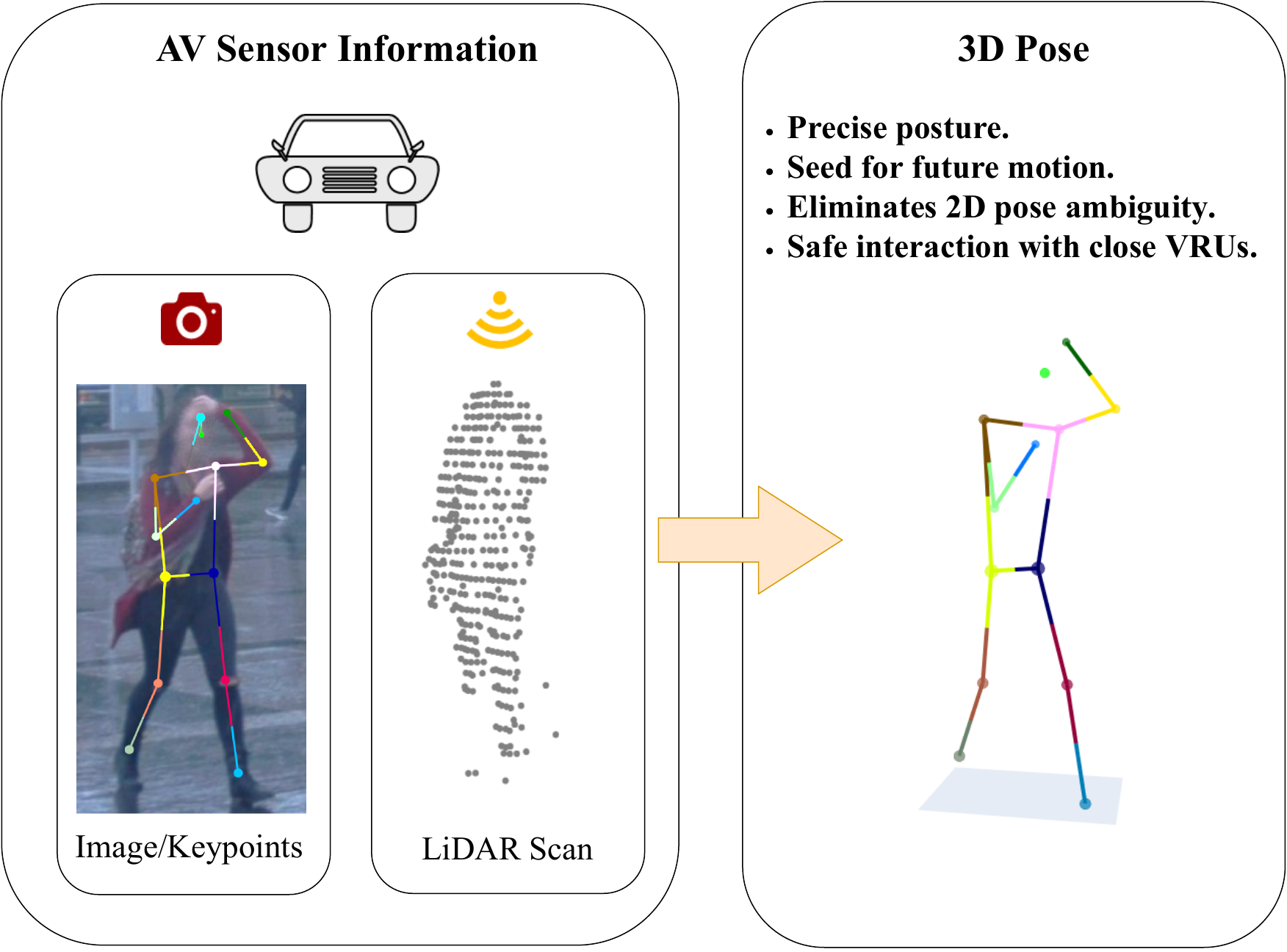}}
\caption{Camera and LiDAR are these days common sensors for autonomous vehicles. However, inferring  the 3D pose of pedestrians and cyclists from 2D joints as well as from a sparse LiDAR scan is a non-trivial, yet needed for many safety critical features such as motion forecasting, trajectory prediction as well as intention or gesture recognition. Furthermore, labeling human 3D poses in uncontrolled outdoor scenarios is non-trivial and an error prone task by itself. Therefore, reconstruction in a weakly supervised manner is preferable.}
\label{fig_0}
\end{figure}

\noindent users (VRU) and the ability to predict their future actions and trajectories\cite{xiong2019recurrent,kooij2019context}. 

In this work, we propose an effective method to fuse information extracted from RGB images (i.e. 2D joint points) and the LiDAR point cloud to obtain a 3D human body pose estimate of the vulnerable road user. In detail, two independent modality-specific branches process the information and the result of each branch is weighted by a linear activation layer to obtain the final body pose. In our ablation studies, we show that each branch is contributing to the final prediction and the network is able to outperform single modalities by using the proposed fusion technique.
Using an off-the-shelf 2D joint point extractor \cite{fang2022alphapose}, we build a weakly supervised 3D HPE model based on pseudo labels. We show that our model is able to improve performance compared to the direct evaluation on pseudo labels while being decoupled from LiDAR to image projections during inference.
While the 2D joint point extractor is pretrained on COCO \cite{Lin} keypoint labels, only accurate LiDAR to image projections are required to train a 3D pose model from 3D VRU detections.
Thus, our weakly supervised setting facilitates an efficient domain transfer, circumvents the error-prone and tedious annotation process for 3D body poses in point clouds, and provides an efficient alternative at almost no additional cost. 
      
\begin{table*}[h!]
\caption{3D Human Pose Estimation Datasets for Autonomous Driving.}
    \centering
    \begin{tabular}{c|cccccc}
    \hline
    Dataset & Frames & Num. 3D poses. & 3D label Type & LiDAR & Camera & AV usable \\
    \hline
    H36M \cite{Ionescu2014} &  $\sim$ 3.6M &  $\sim$ 3.6M & mocap & \xmark & multi-view & \xmark \\
    
    PedX  \cite{Kim} & $\sim$ 5K  & 626 & mocap & \cmark & stereo & \cmark \\

    WOD \cite{Zheng} &  $\sim$ 406K & $\sim$ 10.6K & point cloud & \cmark & mono & \cmark\\
    \hline
    \end{tabular}
    \label{tab:compare_3D_pose_methods}
\end{table*}

The main contributions of our work are threefold: 
\begin{itemize}
    \item We propose an efficient, multi-modal architecture for 3D human pose estimation in the AV context. Our high-level sensor fusion approach operates even if one modality is not available and can be applied in a supervised and a weakly supervised setting. 
  
    \item The weakly supervised setting is based on pseudo labels similar to \cite{Zheng}, but with a more flexible label generation strategy. To the best of our knowledge, our approach is the first to be trained in the AV context without any 2D~/~3D keypoint labels on the target dataset. Only accurate LiDAR to image projections are required, avoiding an error-prone and tedious annotation process in point clouds.
  
    \item We perform extensive experiments on the Waymo Open Dataset v1.3.2 (WOD)\cite{Sun} and demonstrate that our approach outperforms state-of-the-art results \cite{Zheng} by up to $\sim$~13\% on the WOD in the weakly supervised setting and achieves state-of-the-art results\cite{zanfir2022hum3dil} in the purely supervised setting.
\end{itemize}

The remainder of this paper is organized as follows. In Section 2, we introduce the related work to human pose estimation. In Section 3, we illustrate the details of the fully- and weakly-supervised approach. Experimental results and discussion are presented in Section 4.

\section{Related Work}
HPE is an important problem in computer vision. Previous works typically relied on graphical models and multi-view geometry. With the advent of deep learning, many monocular pose estimation approaches have emerged. Below, we present the relevant works
on learning-based 3D HPE.\\

\textbf{Keypoint Lifting.} Thanks to the availability of large-scale datasets \cite{Zheng,Kim,braun2019eurocity}  and the advent of 2D HPE approaches \cite{wang2021urbanpose,braun2021simple,kumar2021vru}, human pose estimation experienced a huge success in automated driving in the last years. Capitalizing this  success, many approaches propose to \textit{lift} the estimated 2D poses into 3D space. More precisely, in the first stage, a pretrained detector is used to obtain the 2D skeletal representation in the image plane. Subsequently, a lifting network infers the 3D pose conditioned on the low-dimensional input. Martinez \textit{et al.} \cite{Martinez2017} proposed an effective residual network, indicating that a large error source of direct estimation methods stems from their visual analysis. Pavallo \textit{et al.} \cite{pavllo20193d} show that human 3D body pose can be effectively learned from video sequences with a fully temporal convolutional model based on dilated temporal convolutions over 2D poses. More recent methods rely on graph convolutional networks \cite{Ci, Zhao, Zou} or transformers \cite{ZhengTransformer, LiTransformer, LiTransformer2 } that reconstruct the human pose from video sequences. Despite the impressive results, these approaches
require still ground-truth 3D body poses. In this work, we show how to reach similar performance without requiring 3D annotations. \\

\textbf{Learning from Point Clouds.}
As a major challenge in 3D HPE, the depth ambiguity of cameras can be alleviated by additional sensors.
While most indoor approaches utilize RGB-D cameras, LiDAR has become the de facto standard system for depth perception in outdoor environments. PointNet \cite{PointNet} is one of the first and most well-known examples for point cloud-based classification and segmentation. PointNet++ \cite{PointNet++} builds on that success with hierarchical feature learning. By using PointNet recursively, it is able to learn local features with increasing context scale through exploiting metric spatial distances. Lately, Zhang \textit{et al.} \cite{Zhang1} proposed a weakly supervised adversarial framework for RGB-D images that was extended to image sequences by means of a density-guided attention-based differentiable point cloud sampling method \cite{Zhang2}. However, their approach is tailor-made for dense point clouds as known from indoor applications and not designed to work with sparse LiDAR data. \\

\textbf{3D HPE for Autonomous Driving.} Kim et al. \cite{Kim} did some pioneer work by introducing the PedX dataset which consists of more than 5000 pairs of stereo images and 2500 LiDAR scans. A 3D model fitting algorithm constrained on different modalities was utilized to obtain automatic 3D labeling of pedestrians. Instead of training a deep neural network and learning the 3D body pose from given inputs as commonly done in pose estimation, the authors minimize multiple energy terms constrained on camera labels and the LiDAR data.
In comparison to Kim \textit{et al.} \cite{Kim}, we propose a learning-based framework for pose estimation which is able to operate in a supervised and weakly supervised setup. Additionally, our approach is able to work with a monocular camera and without 2D segmentation labels. While Zheng \textit{et al.} \cite{Zheng} utilize manual 2D keypoint annotations as seed for the pseudo label generation, we introduce a method that can be trained without any keypoint labels but instead uses an off-the-shelf joint point extractor to produce reasonable 2D poses.
This allows us to train on all datasets containing 3D / 2D detections and LiDAR to image projections, making it almost cost-free to obtain a 3D HPE model if appropriate data is available. While finalizing this work, Zanfir \textit{et al.} \cite{zanfir2022hum3dil} proposed their transformer-based semi-supervised human pose estimation method for autonomous vehicles. However, their method is only able to operate in a semi-supervised manner and is not applicable to datasets without any 3D labels as commonly available in the AV domain. In
Tab. \ref{tab:compare_3D_pose_methods} we give a characteristic comparison of the previously discussed datasets.

\section{Method}

\subsection{Problem Formulation}
We define the human body pose by a fixed number of keypoints $\bold{Y} = [\textbf{y}_1,\textbf{y}_2, ..., \textbf{y}_{N_J}] \in  \mathbb{R}^{N_J \times 3}$ in three-dimensional space. Based on the given 2D keypoint input $\bold{X} = [\textbf{x}_1,\textbf{x}_2, ..., \textbf{x}_{N_J}]\in  \mathbb{R}^{N_J \times 2}$ from the image and the point cloud data $\textbf{P} = \left[\textbf{p}_1, \textbf{p}_2, ..., \textbf{p}_N\right] \in \mathbb{R}^{N \times 3}$, our goal is to infer the pose $\bold{Y}$ by means of the functional mapping

\begin{equation}
    \textbf{Y} = f_{\theta}\left(\textbf{P}, \textbf{X}\right),
\end{equation}

\noindent
where $f$ is approximated by a neural network with a parameter set $\theta$. Additionally, we assume the extrinsic $\textbf{A}_{ext}$, intrinsics $\textbf{A}_{int}$, and the center of the 3D bounding box $\textbf{bb}_{3D}$ to be given for each sample. Using this setup, the projection onto the image plane can be calculated for every point in space.

\subsection{Supervised Approach}
\begin{figure}[h]
\centerline{\includegraphics[width=0.45\textwidth]{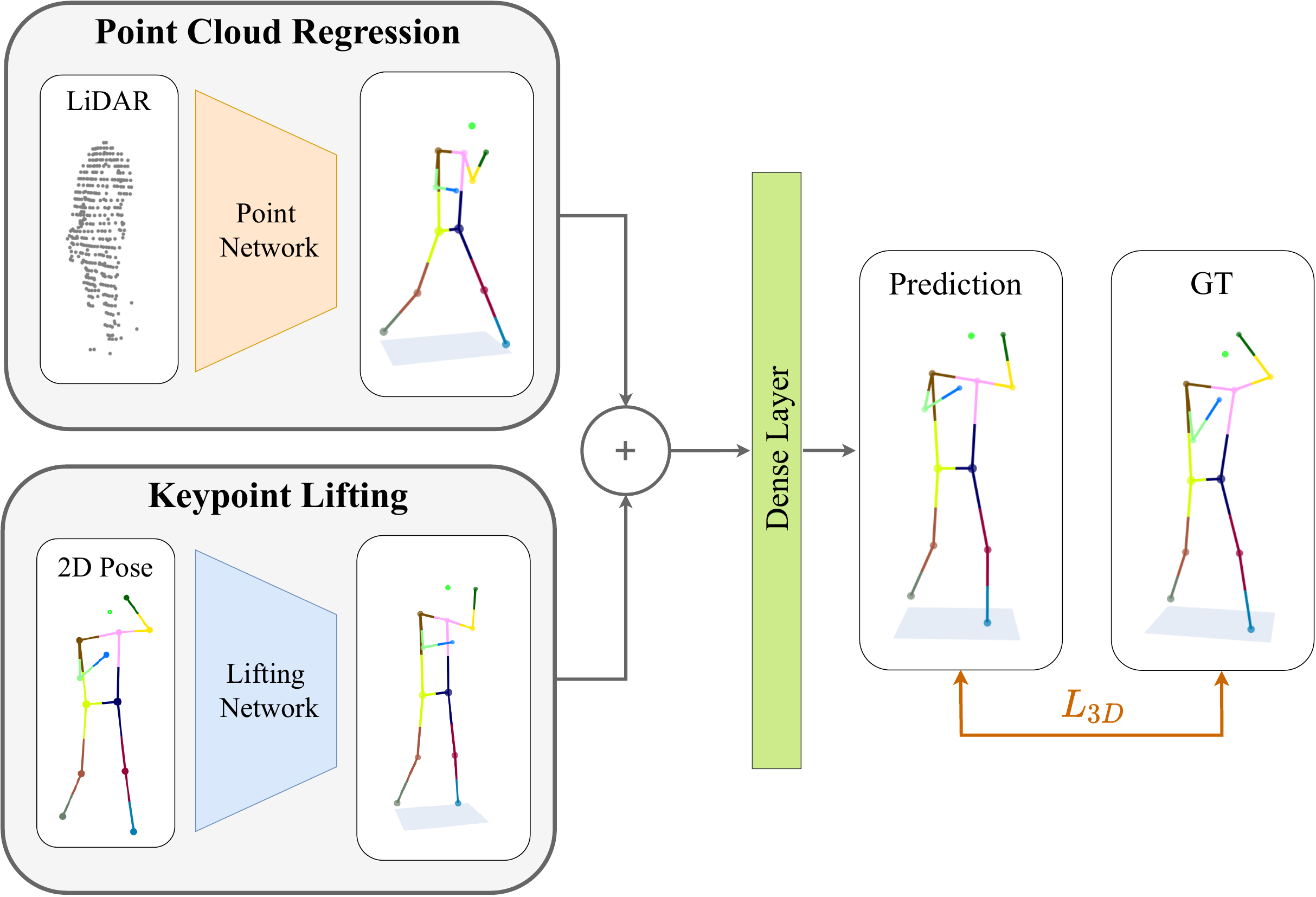}}
\caption{Our proposed supervised framework takes sparse keypoint and point cloud input and processes the data separately. Finally, the branch-wise predictions, which can be considered feature embeddings, are fused by a single dense layer that weights the different contributions. If ground truth information is available, the approach can be trained in a supervised manner.}
\label{fig_1}
\end{figure}

For supervised learning, the 3D ground truth labels are available during training. 
The keypoint branch adopts a similar architecture as Martinez \textit{et al.} \cite{Martinez2017}, using dense layers with dropout, RELU-activation, and residual connections. For the point network, PointNet from Qi \textit{et al.} \cite{PointNet} is adapted to perform a regression instead of a classification. Therefore, we remove the segmentation branch of the original implementation and replace the softmax classification with three fully connected layers that map the global feature vector to the desired output format.     
Finally, the different predictions are fused by an additional dense layer that weights the contributions of each branch.
An overview of the proposed architecture is given in Figure~\ref{fig_1}. \

\subsection{Weakly Supervised Approach}
\label{subseq:ws}

\begin{figure*}[htbp]
\centering
\includegraphics[width=0.7\textwidth]{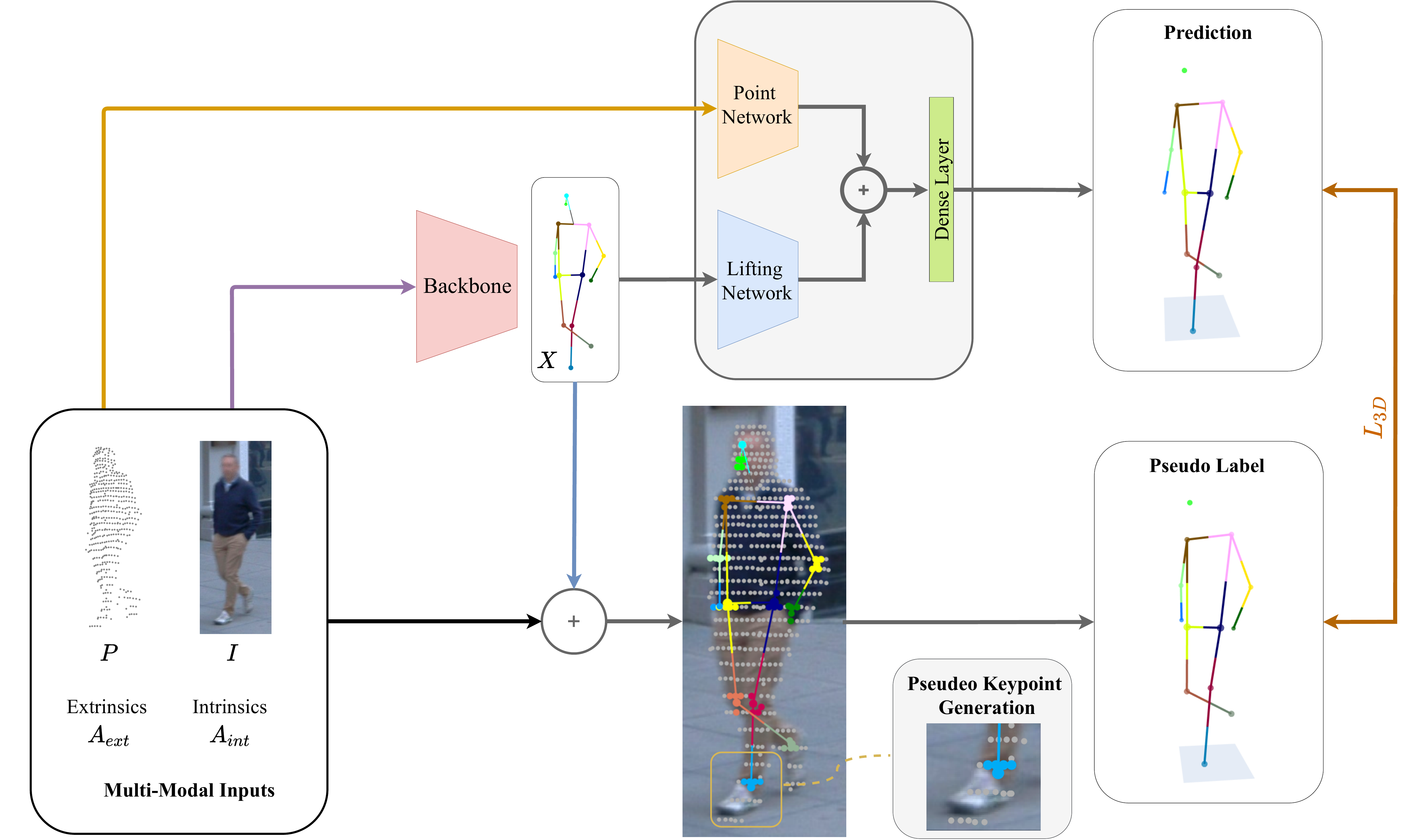}
\caption{\textbf{Weakly supervised approach}. Camera inputs $\textbf{I}$ are processed by an off-the-shelf pose extractor to generate the sparse 2D keypoint representation $\textbf{X}$. The LiDAR point cloud $\textbf{P}$ gets processed by a point network and the 2D pose is lifted into 3D space. Both intermediate representations are fused by a dense layer to obtain the final 3D pose estimation. In parallel,
LiDAR points are projected onto the image plane by using the intrinsics $\textbf{A}_{int}$ and extrinsics $\textbf{A}_{ext}$. Finally, pseudo labels are created by selecting appropriate LiDAR points that are close to the 2D joints, following the weighting mechanism introduced in \ref{subseq:ws}.}
\label{fig_2}
\end{figure*}

While supervised methods are straightforward to train, data collection and annotation is a demanding and error-prone case for 3D HPE applications in outdoor scenarios. Many popular datasets contain LiDAR data and 3D bounding box annotations \cite{nuScenes, ECP25, kitti}, but lack 3D keypoint labels. Similar to Zheng \textit{et al.} \cite{Zheng}, we introduce a weakly supervised pipeline that uses an off-the-shelf 2D joint point extractor (we used Alpha Pose\cite{fang2022alphapose}) and accurate LiDAR to image projections. Figure~\ref{fig_2} shows the weakly supervised learning method.
In the first step, the projected LiDAR points closest to the 2D keypoint estimation are selected. Afterward, these points are weighted to create a pseudo label that serves as ground truth during training. The contribution of each LiDAR point in space is determined by its weighting factor $\gamma$.

\begin{equation}
    \tilde{\textbf{y}}_{k}^{(3)} = \sum_{m=1}^{N_m} \gamma_{mk} \textbf{p}_m^{(3)}\\
\end{equation}

$N_m$ denotes the set of LiDAR points found in the neighborhood of the 2D keypoint and $\tilde{\textbf{y}}_{k}^{(3)}$ is the $k^{th}$ pseudo label in 3D space. We adopt the convention of using a superscript to indicate whether a point is taken from 2D or 3D.
While recent methods perform the weighting of each selected LiDAR point in the image space \cite{Zheng}, we argue that the contribution should rather be calculated in the point cloud. Therefore, we introduce a novel mechanism that lifts the weighting into 3D space by using the distance to the average of the selected LiDAR points.  

\begin{align}
    \gamma_{mk} &=\frac{\exp\left(-\lVert \textbf{p}_m^{(3)} -\Bar{\textbf{p}}_k^{(3)}\rVert_2\right)}{\sum_{n \in N_m}\exp\left(-\lVert \textbf{p}_n^{(3)} -\Bar{\textbf{p}}_k^{(3)}\rVert_2\right)}\\
    \Bar{\textbf{p}}_k^{(3)} &= \frac{1}{|N_m|}\sum_{n \in N_m} \textbf{p}_{nk}^{(3)}
\end{align}

 We prove that our method is efficient by evaluating the pseudo labels directly on ground truth annotations as explained in section \ref{subsubseq:ExpWeakly}.


\section{Experiments}
\label{seq:experiments}
\subsection{Dataset}
\label{subsec:dataset}
We use the publicly available WOD \cite{Sun} for all experiments. The WOD is a large-scale perception dataset that contains roughly 10.6K VRU point cloud labels in the training and validation set. For both supervised and weakly supervised approaches, we utilize the data from the official training dataset to tune our models. The official WOD validation set serves for testing. Since our models rely on 2D information, we remove all 3D labels without a 2D~/~3D label correspondence from the original data ($\sim$ 5K). Additionally, samples that contain less than 75 LiDAR points or less than seven labeled 2D keypoints are omitted.
To ensure that the weakly supervised approach works as expected, LiDAR to image projections have to be accurate. While this holds for most cases, some projections fall into an overlap of different cameras. Projections are only returned for one image, therefore, samples are selected if at least 75\% of the points from the 3D bounding box are projected to the specified camera. In total, our data cleaning removes 10.4\% from the training and 8.2\% from the validation data. This leads to the final supervised dataset used in our experiments, which contains 4169 (83.4\%) samples for training and 831 (16.6\%) for testing. For the weakly supervised setup, the training dataset can be enlarged because no ground truth information is required. This enables us to train on roughly 226K samples automatically annotated through \cite{fang2022alphapose}. The selection of the weakly supervised examples follows the supervised approach. However, instead of seven labeled joints, we propose to make use of the confidence score of the 2D joint extractor backbone (details given in \ref{subsec:training}). We use the same keypoint description as in \cite{Zheng}, leading to 13 unique body joints.
 Note that we consider our filtering mechanism as rather conservative, Zheng \textit{et al.} \cite{Zheng} sub-sample the input point cloud to 256 points while our approach is able to operate with less dense point clouds.

\subsection{Data Preparation}

\textbf{2D keypoints} are labeled in the original image space. Due to the projection property of images, the size of a person is directly related to its distance from the camera. To bypass the scale ambiguity of 2D annotations, all samples are normalized by their height before training. More specifically, each sample is mapped to a height from $\left[-1,+1\right]$ and a width that keeps the original aspect ratio, as introduced in \cite{pavllo20193d}.
For each 2D keypoint vector $\textbf{x}_i$ its normalization $\bar{\textbf{x}}_i$ can be calculated using
\begin{equation}
     \bar{\textbf{x}}_i =  \frac{2\cdot\textbf{x}_i}{h} - \begin{bmatrix} w/h \\ 1 \\ \end{bmatrix}
\end{equation}

where $w$ is the width and $h$ is the height of the 2D pose.

\textbf{3D keypoints} and the LiDAR \textbf{point cloud} are provided in the original vehicle-centered system. By using the 4x4 "extrinsics" transformation matrix from the WOD, the coordinate system is transformed to the sensor coordinates of the camera that captures the vulnerable road user.

Recent works suggest predicting the human pose based on a hip-centered coordinate system \cite{Martinez2017, Drover2018, Kocabas2019, Bouazizi2021}. While this setup ensures a consistent description of all body parts and forces them to lie within certain ranges, the center between the hips is typically unknown when operating in an uncontrolled outdoor environment and cannot easily be inferred from the 3D bounding box detection of a person.
This leads us to a description of the keypoints with respect to the 3D bounding box center.

\begin{equation}
      \hat{\textbf{p}}_i = \textbf{A}_{ext} \cdot \left( \textbf{p}_i - \textbf{bb}_{3D}\right)
\end{equation}

Note that this coordinate system heavily depends on the accuracy of the bounding box detection and contains less implicit information about the location of mostly static body parts (hips or shoulders). Hence, pose predictions with respect to the bounding box-centered system may be considered the more difficult task.

\subsection{Implementation Details}
\label{subseq:implementations_details}

Many 2D pose extractors can be applied for real-time applications \cite{fang2022alphapose}. Our lightweight weakly supervised method builds on top of these modules to infer the human pose in 3D space. A single forward pass on our complete architecture can be done in  $\sim 4.5$ milliseconds on a NVIDIA GeForce RTX 2080 Ti GPU.
This demonstrates the efficiency of our approach.
The supervised model is trained for 250 epochs using the ADAM optimizer \cite{adam} with a learning rate of $5 \times 10^{-4}$. The weakly supervised model is trained for 25 epochs and a learning rate of $5 \times 10^{-4}$. Additionally, we replace the image keypoint labels with predictions from the 2D pose generation backbone and set the confidence threshold $t=0.8$. We use ADAM optimizer \cite{adam} and an exponential learning rate decay. The lifting and the point cloud branch are trained jointly. Our model is implemented in PyTorch and a random seed is set to 42.

\textbf{Point network:} We sample the data to a fixed size of 512 points before feeding it into the point network module. A dropout of 0.4 is applied prior to the final regression layer.

\textbf{Lifting network:} The lifting model consists of four residual blocks with 512 neurons as introduced by Martinez \textit{et al.}. We apply a dropout of 0.1 after each layer.

\subsection{Metrics}
The mean per joint position error (MPJPE) is the most commonly applied metric for 3D human pose estimation \cite{Ionescu2014, Kim}, as well as for motion forecasting  \cite{Cao2020, onHuman, Wang, Mao,ijcai2022p111}.
Let $\textbf{y}_i$ describe the ground truth location of the $i^{th}$ joint in 3D space and $\hat{\mathbf{y}}_i$ the corresponding prediction.
Then, the MPJPE is computed as

\begin{equation}
    E_{mpjpe} = \frac{1}{N_J} \sum_{i=1}^{N_J} {\left \lVert \hat{\textbf{y}}_i - \textbf{y}_i \right \rVert}_2
\end{equation}

\noindent
with $N_J$ as the number of joints in the skeletal body representation. In other words, the MPJPE describes the mean distance between the predicted and ground truth joint location and quantifies the error of the 3D coordinate predictions in a metric unit. During training, the MPJPE serves as the loss measure between predicted and ground truth pose.

\subsection{Training}
\label{subsec:training}

To avoid incorrect supervision during weakly supervised training, image joint points need to be reliable. A pseudo label is only created if an image joint point surpasses the confidence threshold $t$ of the pose extraction backbone. Additionally, we emphasize on the certainty of the pose extraction module by weighting the loss (position error) of each joint based on the confidence score $c$. This forces reliable joints to have a larger impact on the overall loss.

\begin{equation}
     \beta_i = \exp \left(1-\frac{1}{c_i^2} \right) 
\end{equation}

 The loss in terms of Mean Per Joint Position Error (MPJPE) per sample is then given by:
\begin{equation}\label{input_loss}
\mathcal{L}_{3D} =  \frac{1}{N_J} \sum_{i=1}^{N_J} \beta_{i} {\left \lVert \hat{\textbf{y}}_i - \textbf{y}_i \right \rVert}_2
\end{equation}

\subsection{Qualitative Results}
\begin{figure*}[htbp]
\centering
\includegraphics[width=0.7\textwidth]{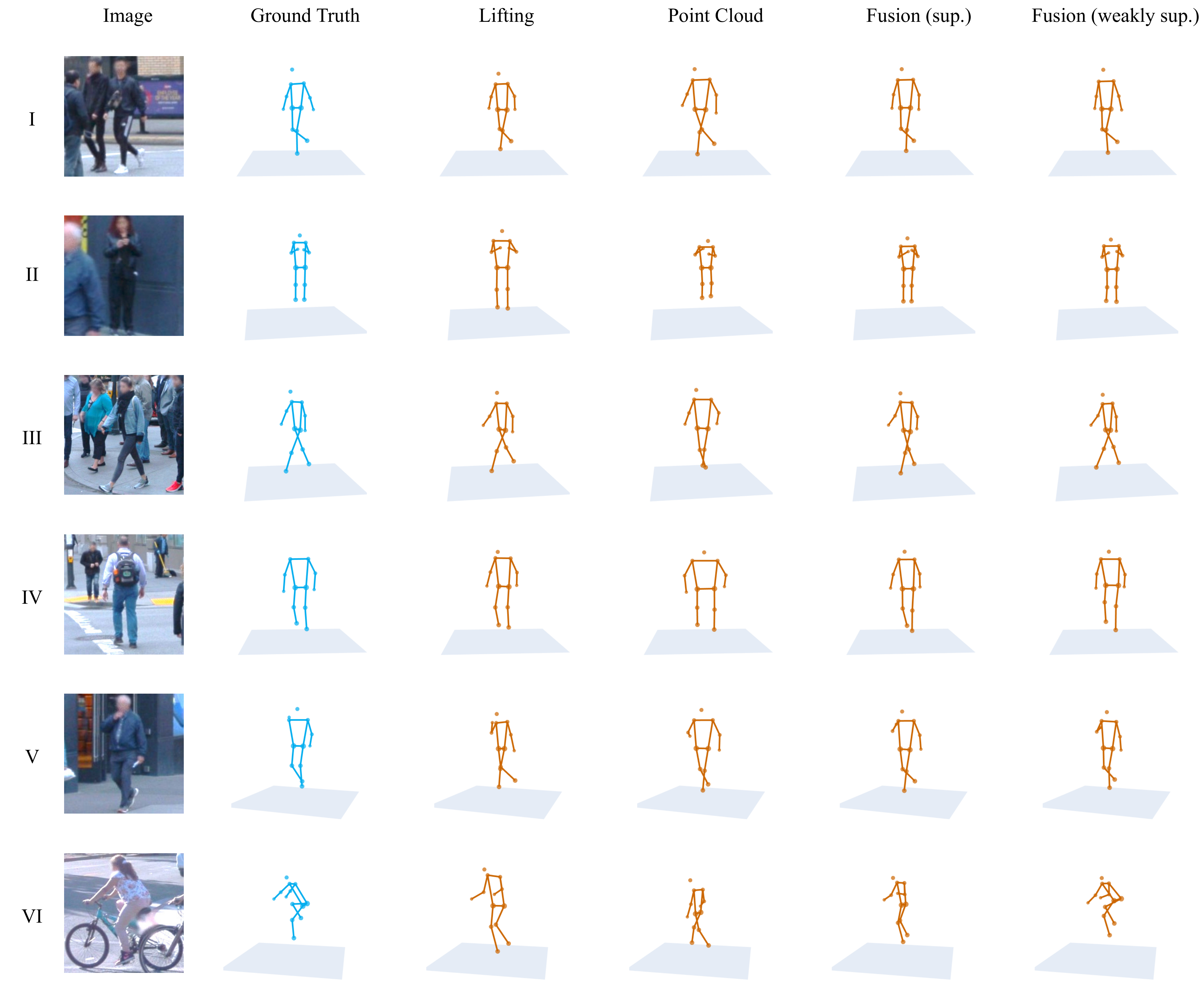}
\caption{Qualitative results of the different approaches. From left to right \textit{image crop}, \textit{ground truth}, \textit{keypoint lifting}, \textit{point cloud regression}, \textit{supervised fusion}, \textit{weakly supervised fusion}. Note that the ground truth labels often miss certain joints which are hard or impossible to label. While the supervised fusion approach is able to accurately recover the pedestrian's pose in many common cases, reconstruction fails for highly unusual cases which are not sufficiently covered by the relatively small training dataset. As an example compare the cyclist pose in row VI: The model trained with the weakly supervised setting has seen a larger amount of data and thus provides a more complete posture approximation.}
\label{fig_3}
\end{figure*}

We show some qualitative results from the WOD for all models trained in Figure~\ref{fig_3}. While the lifting network can generally reconstruct the pose semantics, it often lags pose size and orientation due to the projection ambiguity. In comparison to that, the point network is able to determine the direction and scale of the observed pedestrian but has difficulties with joints held close before the body. An example is shown in picture V with a pedestrian touching his neck. While the lifting network is able to capture the position of the wrist, the point network is unable to differentiate between the wrist and the chest anymore. For keypoint lifting, the poses itself often seem reasonable, however, the orientation in 3D space is problematic and the final predictions have slight rotations with respect to the actual target. In both cases, the fusion network combines the information from the different data sources and provides the most accurate prediction. Finally, picture IV provides an example of a rather rare case in the supervised training dataset, a person riding a bike. It can be seen that in this case the weakly supervised model gives the most accurate prediction and the supervised methods fail to reproduce the pose. This behavior can be explained by the amount of data utilized for training of supervised and weakly supervised methods. Since the weakly supervised method is trained on many samples it includes more uncommon poses, however, the labels themselves  are noisier which leads to the generally worse performance observed.

\subsection{Quantitative Results}
\label{subseq:Qualitative}
\subsubsection{Supervised Network Ablations}
We report the results of single modality models (lifting, point cloud) as well as the fusion scores in Table \ref{tab:results}. The fusion model achieves a $\sim 31\%$ relative improvement over the keypoint-only model from Martinez \textit{et al.} \cite{Martinez2017} and a $\sim 24\%$ improvement compared to the implemented point network. Our approach is able to achieve similar results as the supervised variant of HUM3DIL \cite{zanfir2022hum3dil} recently proposed. These ablations highlight the importance of the different modalities and the unique information carried with them. While the point cloud data itself is able to recover good estimates, the semantic information from the pose representation in the image space is able to refine the prediction.

\begin{table}[]
\caption{Comparison of Supervised 3D-HPE Methods on the WOD.}
    \centering
    \begin{threeparttable}

    \begin{tabular}{cccc}
        \hline
         Methods  & Keypoints / Image & LiDAR & MPJPE [cm] \\
         \hline
         Lifting \cite{Martinez2017}$^{\mathrm{\dag}}$ & Keypoints & & 12.52 \\
         PointNet \cite{PointNet}$^{\mathrm{\dag}}$ &  & \cmark&  11.22\\
         HUM3DIL w/$\lambda=0$\cite{zanfir2022hum3dil}$^{\mathrm{\star}}$ & Image & \cmark & 8.62\\  
         \textit{\textbf{Ours} (supervised fusion)}  & Keypoints  & \cmark & \textbf{8.58} \\
        \hline
        
         
    \end{tabular}
    
    \begin{tablenotes}
    \item $^{\mathrm{\dag}}$ Adapted form the original implementations as described in \ref{subseq:implementations_details}.
    \item $^{\mathrm{\star}}$ Results are provided on a non-public test split, i.e. 50\% of original WOD validation set (random split not available). Note that we show our results on the WOD validation set modified as descriebd in \ref{subsec:dataset}.
    \end{tablenotes}
    \end{threeparttable}
    \label{tab:results}
\end{table}

\subsubsection{Weakly Supervised Learning}
\label{subsubseq:ExpWeakly}
To demonstrate that lifting the pseudo label generation into 3D space is effective, we report the results of a direct evaluation on pseudo labels. In detail, we implement our proposed method as well as the pseudo label generation approach introduced by Zheng \textit{et al.} \cite{Zheng} and compute the MPJPE of the generated label on the testing dataset. Furthermore, we train a model in a weakly supervised setting as introduced in \ref{subseq:ws}. Table \ref{tab:results_weak} contains the obtained results. Our proposed label generation strategy is outperforming existing methods by $2.78$ centimeters ($\sim$~23\%) on the test dataset. We explain the performance gain through the fact that close points in 2D space are not necessarily close in 3D. In other words, by weighting the contributions of each LiDAR point with respect to the mean in 3D, we prevent heavily relying on a projected point that might contain erroneous information about the actual location of the joint point. Weighting with respect to the mean balances the contributions and outliers have less influence. Our results illustrate that models trained on pseudo labels are able to outperform a direct pseudo-label evaluation through weighting different joints based on the confidence score of the 2D pose estimation module. Additionally, the learning-based approach is decoupled from LiDAR to image projections during testing and does not rely on accurate calibrations between the different modalities. We are unable to directly compare our approach with previous work \cite{Zheng}, since their training was done on a non-public internal dataset and not further specified test samples from the WOD.
However, our model achieves a  $\sim$ 13\% relative improvement evaluated on a similar dataset with less dense point clouds as well as noisy keypoint inputs from a joint extraction backbone.


\begin{table}[]
\caption{MPJPE for Weakly Supervised 3D-HPE Models and Pseudo Labels.}
    \centering
    \begin{threeparttable}
    \begin{tabular}{ccc}
        \hline
         Methods  &  Pseudo Labels [cm] & Model [cm] \\
         \hline
         Zheng \textit{et al.}\cite{Zheng} $^{\mathrm{\star}}$ & 12.14 & 10.32\\
         \textbf{\textit{Ours}} &  \textbf{9.36} &  \textbf{9.01} \\
         \hline
                
    \end{tabular}
    \begin{tablenotes}
    \item $\mathrm{\star}$ 
         Results are provided using a non-public internal training set and custom (not specified) test samples. Further, Zheng \textit{et al.} created pseudo labels based on 2D pose ground truth data while we employ a noisy 2D pose extractor to create pseudo labels.
    \end{tablenotes}
    \end{threeparttable}
    \label{tab:results_weak}
\end{table}

\section{Conclusion}
3D human pose estimation for autonomous vehicles differs from classical indoor applications by the sensor setup as well as the unpredictable environment. In this work, we presented a novel method for fusing widely used 2D image joint points and LiDAR data to obtain accurate 3D pose predictions in uncontrolled outdoor scenarios. Our qualitative observations as well as the quantitative results imply that both modalities carry unique information that can be fused on a high level to enhance the final pose estimation. In addition, we show that lifting the pseudo label generation from previous work into the 3D domain by using an average of the selected points increases the robustness of the label and improves the accuracy. Our weakly supervised learning approach is able to train on a target dataset without any 2D~/~3D keypoint labels and solely depends on pedestrian detections and LiDAR to image projections. We outperform the state-of-the-art by up to $\sim$13\% on the Waymo Open Dataset in the weakly supervised setting and achieve state-of-the-art results in the supervised setting.


\section*{Acknowledgment}
The research leading to these results is funded by the German Federal Ministry for Economic Affairs and Energy within the project “KI Delta Learning” (Förderkennzeichen 19A19013A). The authors would like to thank the consortium for the successful cooperation. 

 \input{bib.short.def}
\bibliographystyle{IEEEtran}
\bibliography{bib.bib}
\end{document}